\documentclass[conference]{IEEEtran}
\IEEEoverridecommandlockouts

\usepackage{cite}
\usepackage{amsmath,amssymb,amsfonts}
\usepackage{algorithmic}
\usepackage{graphicx}
\usepackage{textcomp}
\usepackage{xcolor}
\usepackage{booktabs}
\usepackage{multirow}
\usepackage{multicol}
\usepackage{hyperref}
\usepackage{url}
\usepackage[most]{tcolorbox} 
\def\BibTeX{{\rm B\kern-.05em{\sc i\kern-.025em b}\kern-.08em
    T\kern-.1667em\lower.7ex\hbox{E}\kern-.125emX}}

\begin{document}

\title{EQ-5D Classification Using Biomedical Entity-Enriched Pre-trained Language Models and Multiple Instance Learning}

\author{\IEEEauthorblockN{Zhyar Rzgar K Rostam\IEEEauthorrefmark{1}\IEEEauthorrefmark{2}, and G\'{a}bor Kert\'{e}sz\IEEEauthorrefmark{2}}  
\IEEEauthorblockA{\IEEEauthorrefmark{1}Doctoral School of Applied Informatics and Applied Mathematics, Obuda University, Budapest, Hungary \\}  
\IEEEauthorblockA{\IEEEauthorrefmark{2}John von Neumann Faculty of Informatics, Obuda University, Budapest, Hungary \\
Emails: \{kwekha.rostam.zhyar, kertesz.gabor\}@nik.uni-obuda.hu}  
}

\maketitle

\begin{abstract}
The EQ-5D (EuroQol 5-Dimensions) is a standardized instrument for the evaluation of health-related quality of life. In health economics, systematic literature reviews (SLRs) depend on the correct identification of publications that use the EQ-5D, but manual screening of large volumes of scientific literature is time-consuming, error-prone, and inconsistent. In this study, we investigate fine-tuning of general-purpose (BERT) and domain-specific (SciBERT, BioBERT) pre-trained language models (PLMs), enriched with biomedical entity information extracted through scispaCy models for each statement, to improve EQ-5D detection from abstracts. We conduct nine experimental setups, including combining three scispaCy models with three PLMs, and evaluate their performance at both the sentence and study levels. Furthermore, we explore a Multiple Instance Learning (MIL) approach with attention pooling to aggregate sentence-level information into study-level predictions, where each abstract is represented as a bag of enriched sentences (by scispaCy). The findings indicate consistent improvements in F1-scores (reaching 0.82) and nearly perfect recall at the study-level, significantly exceeding classical bag-of-words baselines and recently reported PLM baselines. These results show that entity enrichment significantly improves domain adaptation and model generalization, enabling more accurate automated screening in systematic reviews.

\end{abstract}


\section{Introduction}
\label{introduction}
Systematic literature reviews (SLRs) are one of the most essential and composite strategies for evidence synthesis, especially regarding health economic and clinical research. However, the recent increase in the number of scientific manuscripts published annually leads to traditional titles, and abstract screening is time-consuming, error-prone, and inconsistent~\cite{ahanger2022novel, app13063594}. This is more observable when complex inclusion criteria are applied, such as when evaluating studies that use and state EQ-5D as a recognized standardized measure of health-related quality of life~\cite{kertesz2024eq5d, zah2022paying}.


Recent developments in natural language processing (NLP)~\cite{peng2021survey, naseem2022benchmarking} and deep learning (DL)~\cite{jiao2023brief, alimova2021cross} have enabled automated text classification. Pre-trained language models (PLMs) that utilize Transformer as their architecture have achieved remarkable success in most of the NLP tasks, such as sentiment analysis~\cite{alimova2021cross, laki2023sentiment}, topic modeling~\cite{kherwa2019topic, lezama2023integrating}, information retrieval, and natural language inference. However, PLMs effectiveness declines in domain-specific tasks due to specialized terminology and imbalanced data distributions~\cite{beltagy2019scibert, lee2020biobert, peng2019transfer}. 

In this study, we evaluate the performance of general-purpose (BERT)~\cite{devlin2018bert} and domain-specific (SciBERT and BioBERT)~\cite{lee2020biobert, beltagy2019scibert} PLMs for SLRs in classifying whether studies mention EQ-5D solely on the abstract. We apply different techniques and strategies, such as enriching abstract statements with three different versions of scispaCy (\texttt{en\_core\_sci\_sm}\footnote{\texttt{en\_core\_sci\_sm}: A lightweight spaCy pipeline for biomedical data.}, 
\texttt{en\_core\_sci\_md}\footnote{\texttt{en\_core\_sci\_md}: A spaCy pipeline with an extended vocabulary and 50k word vectors for biomedical data.}, 
\texttt{en\_core\_sci\_scibert}\footnote{\texttt{en\_core\_sci\_scibert}: A spaCy pipeline with a $\sim$785k vocabulary, using \texttt{allenai/scibert-base} as the transformer model.})
 individually, then feeding the enriched statements to the PLMs for classification. Predictions are computed by averaging the prediction confidence scores according to which sentences in the studies belong to that category. The final prediction is derived by selecting the label with the maximum confidence. Additionally, we investigated a Multiple Instance Learning (MIL) approach with attention pooling to aggregate sentence-level information into study-level predictions, where each abstract is defined as a bag of enriched sentences. The aims of this study are to:

\begin{itemize}
    \item Investigate the impact of entity enrichment using scispaCy (\texttt{en\_core\_sci\_sm}, \texttt{en\_core\_sci\_md}, and \texttt{en\_core\_sci\_scibert}) on fine-tuned PLMs.
    \item Evaluate the effectiveness of general-purpose (BERT) and domain-specific PLMs (SciBERT, BioBERT) for EQ-5D classification in biomedical abstracts.
    \item Compare classification performance between study-level and sentence-level approaches to enhance systematic review automation. 
    \item Benchmark the proposed pipeline against classical and fine-tuned baseline approaches.
    \item Explore a MIL with study-level prediction that uses the attention pool.
\end{itemize}

\section{Related Works}

\subsection{Semi-automated Screening}

Wallace et al.~\cite{wallace2010semi} suggest a method for semi-automating the citation screening through SVM ensembles along with an active learning strategy, like PAL, to deal with imbalanced datasets. With this entire mechanism, their approach gives a reduction in manual screening by up to 50\%, together with maintaining the reviews in multiple datasets.

Cohen et al.~\cite{cohen2006reducing} propose a voting perceptron-based system for classifying citations in systematic drug class reviews. This approach, which was evaluated on 15 annotated reviews, reduced the need for manual screening in most topics by more than 50\% in three cases.

\subsection{Domain-Specific PLMs}


Lee et al.~\cite{lee2020biobert}  present BioBERT as a biomedical PLM fine-tuned on PubMed abstracts and PubMed full-texts. In particular, BioBERT achieves significant improvements over general-purpose models on tasks including named entity recognition (NER), relation extraction (RE), and question answering (QA).

Beltagy et al.~\cite{beltagy2019scibert} introduce SciBERT, a domain-specific PLM specifically designed for scientific text and trained on large scientific corpora. SciBERT achieves remarkable performance superiority over BERT and other general-purpose models in various NLP tasks.

Danilov et al.~\cite{danilov2021classification} utilize PubMedBERT and an ensemble strategy to investigate the classification of 630 PubMed abstracts. The findings from the study indicate that PubMedBERT performs well in the classification of short biomedical texts.

\section{Dataset}
\label{dataset}
In this study, we utilize a dataset derived from Kertész et al.~\cite{kertesz2024eq5d}. The dataset is extracted from PubMed, a publicly accessible database that includes over 36 million citations and abstracts in biomedical and life sciences indexing. The dataset is collected by applying only the EuroQol search without any further filter and search terms, resulting in 15,547 records published between 1990 and 2022. From this collection, Kertész et al.~\cite{kertesz2024eq5d} randomly selected 200 studies, based on the predefined eligibility criteria. Two independent experts review the studies and label the selected studies. Disagreements in reviewers' decisions are resolved through discussion until a final decision is reached. The classification is binary, with labels indicating whether the studies mentioned EQ-5D data or not (``true'' or ``false''). The dataset is composed of 200 labeled studies, each containing a title, abstract, keywords, and labels. All abstracts from the dataset are in English, whereas the full-text of the study may be in another language. Therefore, these studies are considered as negative by the reviewers. The labeled dataset (EQ-5D: 200) contains 121 positive and 79 negative  \footnote{The datasets, models notebooks, and results can be accessed at: \url{https://github.com/ZhyarUoS/Biomedical-Entity-Enrichment-for-EQ-5D.git}}.

\section{Methods}
\subsection{Dataset Preparation and Preprocessing}
\label{data_preperation}
During the experiments, the dataset is divided into training (70\%), testing (30\%), and validation (50\% of the testing set) sets.

In this study, we utilize scispaCy (see Section \ref{introduction}) for segmenting abstracts into sentences. scispaCy is a powerful tool that offers PTMs for processing biomedical and scientific texts for different tasks, such as entity recognition and entity linking. The extracted entities are automatically added to each sentence. For example, a sentence containing biomedical entities is transformed into:



\begin{tcolorbox}[colback=gray!10, colframe=black, boxrule=0.5pt, arc=2pt]
\small
This index was found to be highly correlated with a measure of health (EQ-5D) and wellbeing (global QoL), although some differences were apparent. 

{\footnotesize [ENTS: correlated|ENTITY; measure|ENTITY; health|ENTITY; EQ-5D|ENTITY; wellbeing|ENTITY; global|ENTITY; QoL|ENTITY; apparent|ENTITY]}
\end{tcolorbox}

This enrichment step augments the raw text with domain-specific features for downstream classification.\\
Records with missing abstracts are removed, to ensure compatibility with downstream processing, labels are cast into integers. \emph{RandomSampler} is used in iteration with the training set to build mini batches, which produces random batches per epoch, while for the validation and test sets, it is done deterministically using a \emph{SequentialSampler} for reproducible evaluation. 

\paragraph*{Tokenization and Encoding}
Based on the experimental session, the inputs are tokenized with a compatible tokenizer (BERT, SciBERT, BioBERT). Sequences are truncated or padded to a fixed maximum length (256 tokens), and attention masks distinguish padding from actual tokens. Their outputs are returned as PyTorch tensors. Set up with the \texttt{TensorDataset} class, \texttt{input\_ids}, \texttt{attention\_mask}, and \texttt{labels} fed onto \texttt{DataLoader} instances (with batch size 16) for smooth mini-batch processing.

\subsection{Experimental Setup}
\label{expermintal_setup}
In order to classify EQ-5D text only through studies' abstracts, we evaluate the impact of three segmentation models (see Section~\ref{introduction}) in combination with PLMs (BERT, SciBERT, BioBERT) individually. Additionally, we also utilize MIL approach with attention pooling, where each abstract is represented as a bag of enriched sentences to aggregate sentence-level information into study-level predictions. As a result, we obtain outcomes for nine different configurations for each approach. In each configuration, we follow the same splitting strategy as mentioned in section \ref{data_preperation}. \\ 
The fine-tuning process in both approaches is performed on the training set (after data preparation and preprocessing applied) on the sentence-level, in addition with both enriching sentence with biomedical entities and a bag of enriched sentences through scispaCy models. Table~\ref{tab:training_parameters} presents the fine-tuning configuration details in each setup. AdamW optimizer with four different learning rates is investigated. To prevent overfitting and stabilize the fine-tuning process, a learning rate scheduler with warm-up and early stopping strategy (with patience for 5 epochs) are utilized. The fine-tuning is continued for up to 20 epochs, keeping the best performing model check point based on the validation F1-score.

\begin{table}[tbp]
    \centering
    \caption{Fine-tuning Configuration Parameters}
    \label{tab:training_parameters}
    \begin{tabular}{lr}
        \toprule
        Optimizer                   & AdamW \\
        Epochs (max)                & 20 \\
        Learning rates              & \{2$\times10^{-5}$, 5$\times10^{-6}$, 2$\times10^{-6}$, 1$\times10^{-6}$\} \\
        Scheduler                   & Linear with warm-up \\
        Warm-up steps               & 10\% of total training steps \\
        Early stopping patience     & 5 epochs \\
        Epsilon ($\epsilon$)        & $1\times10^{-8}$ \\
        Maximum sequence length     & 256 tokens \\
        Evaluation metric           & F1-score \\ \hline
        \bottomrule
    \end{tabular}
\end{table}

Evaluation is performed at both the sentence- and the study- levels.  Predictions are aggregated by averaging the prediction confidence scoring from sentences belonging to the study, and deriving the final prediction through selecting the label with the maximum confidence. Through these experimental designs, we are able to provide a comprehensive evaluation of preprocessing scispaCy PTMs and PLMs (BERT, SciBERT, and BioBERT) individually.

Additionally, we investigated MIL with attention pooling, where each abstract is defined as a bag of enriched sentences (through scispaCy) and the model aggregates sentence representations into a study-level prediction, using the same PLMs and scispCy models (see Section~\ref{introduction}).

\section{Results}
In this section, we present the obtained results of both approaches on three different PLMs with three different scispaCy models (see Section~\ref{introduction}). Accuracy, precision, recall, and F1-score are reported for each configuration at both sentence- and study- levels. The sentence-level shows how well predictions can be estimated for individual sentences, whereas the study-level and bag of enrichment sentences assess aggregate predictions across sentences within each abstract in a paper. To ensure the robust and reliable results for both approaches and each of the nine configurations we executed five times, the final reported results reflect the averages of the five independent runs.

\subsection{BERT with \texttt{en\_core\_sci\_sm}}
In this setup, the model achieved moderate performance for both sentence- and study- levels. The average for the sentence-levels are: accuracy 0.69, precision 0.70, recall 0.92, and F1-score 0.79. However, in the study-level the performance is slightly higher in recall by 0.99, which is balanced by a lower precision rate of 0.67, and an overall F1-score of 0.80. These findings indicate that the model is highly sensitive in identifying relevant sentences, while at the study level, the false positive errors increase (see Table~\ref{tab:bert_results}).

\begin{table}[htbp]
\centering

\caption{Performance of BERT Model with \texttt{en\_core\_sci\_sm} Across Five Attempts (Sentence- and Study-Level)}
\scalebox{0.8}
{
\begin{tabular}{c l c c c c}
\hline
\textbf{Attempt} & \textbf{Configuration} & \textbf{Accuracy} & \textbf{Precision} & \textbf{Recall} & \textbf{F1-score} \\ \hline
\multirow{2}{*}{1} & Sentence & 0.71 & 0.73 & 0.85  & 0.79 \\
 & Study    & 0.78 & 0.73 & 1.00  & 0.85 \\ \hline
\multirow{2}{*}{2} & Sentence & 0.66 & 0.65 & 1.00  & 0.79 \\
 & Study    & 0.62 & 0.61 & 1.00  & 0.76 \\ \hline
\multirow{2}{*}{3} & Sentence & 0.71 & 0.77 & 0.77  & 0.77 \\
 & Study    & 0.78 & 0.76 & 0.94  & 0.84 \\ \hline
\multirow{2}{*}{4} & Sentence & 0.67 & 0.66 & 1.00  & 0.79 \\
 & Study    & 0.62 & 0.61 & 1.00  & 0.76 \\ \hline
\multirow{2}{*}{5} & Sentence & 0.70 & 0.68 & 0.98  & 0.80 \\
 & Study    & 0.67 & 0.64 & 1.00  & 0.78 \\ \hline
\textbf{AVG} & Sentence & 0.69 & 0.70 & 0.92  & 0.79 \\
\textbf{AVG} & Study    & 0.69 & 0.67 & 0.99 & 0.80 \\ \hline
\end{tabular}

\label{tab:bert_results}
}
\end{table}

\subsection{SciBERT with \texttt{en\_core\_sci\_sm}}
Compared to BERT, SciBERT with \texttt{en\_core\_sci\_sm} achieved superior sentence-level performance. The average achieved accuracy, precision, recall, and F1-score were 0.70, 0.68, 0.98, and 0.80, respectively. It maintained high recall at the study-level, with accuracy, precision, and F1-score of 0.73, 0.69, and 0.82, respectively. The results indicates that SciBERT is powerful in minimizing false negatives but has a moderate loss of precision (see Table~\ref{tab:scibert_results}).

\begin{table}[htbp]
\centering
\caption{Performance of SciBERT Model with \texttt{en\_core\_sci\_sm} Across Five Attempts (Sentence- and Study-Level)}
\scalebox{0.85}
{
\begin{tabular}{c l c c c c}
\hline
\textbf{Attempt} & \textbf{Configuration} & \textbf{Accuracy} & \textbf{Precision} & \textbf{Recall} & \textbf{F1-score} \\ \hline
\multirow{2}{*}{1} & Sentence & 0.71 & 0.69 & 0.96 & 0.81 \\
                   & Study    & 0.75 & 0.71 & 1.00 & 0.83 \\ \hline
\multirow{2}{*}{2} & Sentence & 0.70 & 0.68 & 0.99 & 0.80 \\
                   & Study    & 0.73 & 0.69 & 1.00 & 0.82 \\ \hline
\multirow{2}{*}{3} & Sentence & 0.69 & 0.68 & 0.99 & 0.80 \\
                   & Study    & 0.72 & 0.68 & 1.00 & 0.81 \\ \hline
\multirow{2}{*}{4} & Sentence & 0.70 & 0.68 & 0.99 & 0.80 \\
                   & Study    & 0.72 & 0.68 & 1.00 & 0.81 \\ \hline
\multirow{2}{*}{5} & Sentence & 0.70 & 0.68 & 0.98 & 0.80 \\
                   & Study    & 0.75 & 0.71 & 1.00 & 0.83 \\ \hline
\textbf{AVG}       & Sentence & 0.70 & 0.68 & 0.98 & 0.80 \\
\textbf{AVG}       & Study    & 0.73 & 0.69 & 1.00 & 0.82 \\ \hline
\end{tabular}
}
\label{tab:scibert_results}
\end{table}

\subsection{BioBERT with \texttt{en\_core\_sci\_sm}}
A combination of BioBERT with \texttt{en\_core\_sci\_sm} also obtained competitive results. For sentence-level results, there was an average accuracy of 0.69, precision of 0.69, recall of 0.93, and F1-score of 0.79. The study-level evaluation achieved higher performance with accuracy of 0.74, a precision of 0.70, a recall of 0.99, and F1-score of 0.82. In a comparison with BERT, BioBERT maintained a better balance between precision and recall at the study-level (see Table~\ref{tab:biobert_results}).

\begin{table}[htbp]
\centering
\caption{Performance of BioBERT Model with \texttt{en\_core\_sci\_sm} Across Five Attempts (Sentence- and Study-Level)}
\scalebox{0.85}
{
\begin{tabular}{c l c c c c}
\hline
\textbf{Attempt} & \textbf{Configuration} & \textbf{Accuracy} & \textbf{Precision} & \textbf{Recall} & \textbf{F1-score} \\ \hline
\multirow{2}{*}{1} & Sentence & 0.71 & 0.74 & 0.83 & 0.78 \\
                   & Study    & 0.80 & 0.75 & 1.00 & 0.86 \\ \hline
\multirow{2}{*}{2} & Sentence & 0.68 & 0.72 & 0.82 & 0.77 \\
                   & Study    & 0.77 & 0.74 & 0.94 & 0.83 \\ \hline
\multirow{2}{*}{3} & Sentence & 0.69 & 0.67 & 1.00 & 0.80 \\
                   & Study    & 0.70 & 0.67 & 1.00 & 0.80 \\ \hline
\multirow{2}{*}{4} & Sentence & 0.69 & 0.67 & 1.00 & 0.80 \\
                   & Study    & 0.72 & 0.68 & 1.00 & 0.81 \\ \hline
\multirow{2}{*}{5} & Sentence & 0.70 & 0.67 & 1.00 & 0.81 \\
                   & Study    & 0.72 & 0.68 & 1.00 & 0.81 \\ \hline
\textbf{AVG}       & Sentence & 0.69 & 0.69 & 0.93 & 0.79 \\
\textbf{AVG}       & Study    & 0.74 & 0.70 & 0.99 & 0.82 \\ \hline
\end{tabular}
}
\label{tab:biobert_results}
\end{table}

\subsection{BERT with \texttt{en\_core\_sci\_md}}
In this setup, BERT performed slightly better on the study-level compared to the \texttt{en\_core\_sci\_sm}. The average of accuracy, precision, recall, and F1-score in the sentence-level configuration was: 0.70, 0.71, 0.89, 0.79, respectively. In the study-level results accuracy reached of 0.73, the precision of 0.70, the recall of 1.0, and the F1-score of 0.82. These findings reveal that this scispaCy model in combination with BERT, improves named entity resolution, leading to better aggregation on the study-level (see Table~\ref{tab:bert_md_results}).

\begin{table}[htbp]
\centering
\caption{Performance of BERT Model with \texttt{en\_core\_sci\_md} Across Five Attempts (Sentence- and Study-Level)}
\scalebox{0.85}
{
\begin{tabular}{c l c c c c}
\hline
\textbf{Attempt} & \textbf{Configuration} & \textbf{Accuracy} & \textbf{Precision} & \textbf{Recall} & \textbf{F1-score} \\ \hline
\multirow{2}{*}{1} & Sentence & 0.70 & 0.74 & 0.82 & 0.78 \\
                   & Study    & 0.78 & 0.73 & 1.00 & 0.85 \\ \hline
\multirow{2}{*}{2} & Sentence & 0.68 & 0.68 & 0.93 & 0.78 \\
                   & Study    & 0.63 & 0.62 & 1.00 & 0.77 \\ \hline
\multirow{2}{*}{3} & Sentence & 0.71 & 0.71 & 0.91 & 0.80 \\
                   & Study    & 0.75 & 0.71 & 1.00 & 0.83 \\ \hline
\multirow{2}{*}{4} & Sentence & 0.69 & 0.71 & 0.87 & 0.78 \\
                   & Study    & 0.75 & 0.71 & 1.00 & 0.83 \\ \hline
\multirow{2}{*}{5} & Sentence & 0.70 & 0.70 & 0.92 & 0.80 \\
                   & Study    & 0.75 & 0.71 & 1.00 & 0.83 \\ \hline
\textbf{AVG}       & Sentence & 0.70 & 0.71 & 0.89 & 0.79 \\
\textbf{AVG}       & Study    & 0.73 & 0.70 & 1.00 & 0.82 \\ \hline
\end{tabular}
}
\label{tab:bert_md_results}
\end{table}

\subsection{SciBERT with \texttt{en\_core\_sci\_md}}
Combining SciBERT with \texttt{en\_core\_sci\_md} showed stable performance with sentence-level accuracy of 0.70, precision 0.70, recall of 0.93, and F1-score of 0.80. At the study-level, the model further achieved an accuracy of 0.72, a precision of 0.68, a recall of 1.0, and F1-score of 0.81. Comparing these achieved performances with the \texttt{en\_core\_sci\_sm} variant, this (\texttt{en\_core\_sci\_md}) scispaCy model demonstrates improved precision, indicating lower false positives while maintaining perfect recall at the study-level (see Table~\ref{tab:scibert_md_results}). 

\begin{table}[htbp]
\centering
\caption{Performance of SciBERT Model with \texttt{en\_core\_sci\_md} Across Five Attempts (Sentence- and Study-Level)}
\scalebox{0.85}
{
\begin{tabular}{c l c c c c}
\hline
\textbf{Attempt} & \textbf{Configuration} & \textbf{Accuracy} & \textbf{Precision} & \textbf{Recall} & \textbf{F1-score} \\ \hline
\multirow{2}{*}{1} & Sentence & 0.71 & 0.72 & 0.88 & 0.79 \\
                   & Study    & 0.77 & 0.72 & 1.00 & 0.84 \\ \hline
\multirow{2}{*}{2} & Sentence & 0.68 & 0.67 & 0.95 & 0.79 \\
                   & Study    & 0.63 & 0.62 & 1.00 & 0.77 \\ \hline
\multirow{2}{*}{3} & Sentence & 0.71 & 0.70 & 0.94 & 0.80 \\
                   & Study    & 0.73 & 0.69 & 1.00 & 0.82 \\ \hline
\multirow{2}{*}{4} & Sentence & 0.71 & 0.70 & 0.94 & 0.80 \\
                   & Study    & 0.73 & 0.69 & 1.00 & 0.82 \\ \hline
\multirow{2}{*}{5} & Sentence & 0.71 & 0.70 & 0.94 & 0.80 \\
                   & Study    & 0.73 & 0.69 & 1.00 & 0.82 \\ \hline
\textbf{AVG}       & Sentence & 0.70 & 0.70 & 0.93 & 0.80 \\
\textbf{AVG}       & Study    & 0.72 & 0.68 & 1.00 & 0.81 \\ \hline
\end{tabular}
}
\label{tab:scibert_md_results}
\end{table}

\subsection{BioBERT with \texttt{en\_core\_sci\_md}}
Combination of BioBERT with \texttt{en\_core\_sci\_md} obtained balanced results. In the sentence-level, it achieved an average accuracy of 0.69, a precision of 0.71, a recall of 0.87, and an F1-score of 0.78. The outcomes at the study-level were stronger, with accuracy of 0.75, precision of 0.71, recall of 0.98, and F1-score of 0.82. This setup shows better performance across all the metrics because its precision between the models is improved with higher recall (see Table~\ref{tab:biobert_md_results}).

\begin{table}[htbp]
\centering
\caption{Performance of BioBERT Model with \texttt{en\_core\_sci\_md} Across Five Attempts (Sentence- and Study-Level)}
\scalebox{0.85}
{
\begin{tabular}{c l c c c c}
\hline
\textbf{Attempt} & \textbf{Configuration} & \textbf{Accuracy} & \textbf{Precision} & \textbf{Recall} & \textbf{F1-score} \\ \hline
\multirow{2}{*}{1} & Sentence & 0.70 & 0.71 & 0.90 & 0.79 \\
                   & Study    & 0.73 & 0.69 & 1.00 & 0.82 \\ \hline
\multirow{2}{*}{2} & Sentence & 0.68 & 0.73 & 0.79 & 0.76 \\
                   & Study    & 0.75 & 0.73 & 0.92 & 0.81 \\ \hline
\multirow{2}{*}{3} & Sentence & 0.69 & 0.71 & 0.86 & 0.78 \\
                   & Study    & 0.75 & 0.71 & 1.00 & 0.83 \\ \hline
\multirow{2}{*}{4} & Sentence & 0.70 & 0.71 & 0.89 & 0.79 \\
                   & Study    & 0.75 & 0.71 & 1.00 & 0.83 \\ \hline
\multirow{2}{*}{5} & Sentence & 0.70 & 0.69 & 0.92 & 0.79 \\
                   & Study    & 0.75 & 0.71 & 1.00 & 0.83 \\ \hline
\textbf{AVG}       & Sentence & 0.69 & 0.71 & 0.87 & 0.78 \\
\textbf{AVG}       & Study    & 0.75 & 0.71 & 0.98 & 0.82 \\ \hline
\end{tabular}
}
\label{tab:biobert_md_results}
\end{table}

\subsection{BERT with \texttt{en\_core\_sci\_scibert}}
By combining BERT with \texttt{en\_core\_sci\_scibert}, at the sentence-level, the setup achieved 0.71, 0.71, 0.91, and 0.80 for accuracy, precision, recall, and F1-score, respectively. At the study-level, this setup achieved an accuracy of 0.73, a precision of 0.69, a recall of 1.00, and an F1-score of 0.82. These findings indicate that SciBERT, as a domain-specific PLM, can provide consistent improvements in recall across studies, ensuring high coverage (see Table~\ref{tab:bert_scibert_results}).

\begin{table}[htbp]
\centering
\caption{Performance of BERT Model \texttt{en\_core\_sci\_scibert} Across Five Attempts (Sentence- and Study-Level)}
\scalebox{0.85}
{
\begin{tabular}{c l c c c c}
\hline
\textbf{Attempt} & \textbf{Configuration} & \textbf{Accuracy} & \textbf{Precision} & \textbf{Recall} & \textbf{F1-score} \\ \hline
\multirow{2}{*}{1} & Sentence & 0.71 & 0.71 & 0.91 & 0.80 \\
                   & Study    & 0.73 & 0.69 & 1.00 & 0.82 \\ \hline
\multirow{2}{*}{2} & Sentence & 0.71 & 0.71 & 0.91 & 0.80 \\
                   & Study    & 0.73 & 0.69 & 1.00 & 0.82 \\ \hline
\multirow{2}{*}{3} & Sentence & 0.71 & 0.72 & 0.90 & 0.80 \\
                   & Study    & 0.73 & 0.69 & 1.00 & 0.82 \\ \hline
\multirow{2}{*}{4} & Sentence & 0.71 & 0.72 & 0.90 & 0.80 \\
                   & Study    & 0.73 & 0.69 & 1.00 & 0.82 \\ \hline
\multirow{2}{*}{5} & Sentence & 0.71 & 0.71 & 0.91 & 0.80 \\
                   & Study    & 0.73 & 0.69 & 1.00 & 0.82 \\ \hline
\textbf{AVG}       & Sentence & 0.71 & 0.71 & 0.91 & 0.80 \\
\textbf{AVG}       & Study    & 0.73 & 0.69 & 1.00 & 0.82 \\ \hline
\end{tabular}
}
\label{tab:bert_scibert_results}
\end{table}

\subsection{SciBERT with \texttt{en\_core\_sci\_scibert}}

Combining SciBERT with \texttt{en\_core\_sci\_scibert} demonstrated slightly lower sentence-level performance compared to both mentioned scispaCy models (\texttt{en\_core\_sci\_sm} and \texttt{en\_core\_sci\_md}), by achieving 0.69, 0.71, 0.87, and 0.78 for accuracy, precision, recall, and F1-score, respectively, while in the study-level this setup achieved better results by 0.73, 0.70, 0.97, and 0.81 for accuracy, precision, recall, and F1-score, respectively. However, recall remained high, precision decreased compared to other settings, and these results show a tendency for false positives (see Table~\ref{tab:scibert_scibert_results}).

\begin{table}[htbp]
\centering
\caption{Performance of SciBERT Model \texttt{en\_core\_sci\_scibert} Across Five Attempts (Sentence- and Study-Level)}
\scalebox{0.85}
{
\begin{tabular}{c l c c c c}
\hline
\textbf{Attempt} & \textbf{Configuration} & \textbf{Accuracy} & \textbf{Precision} & \textbf{Recall} & \textbf{F1-score} \\ \hline
\multirow{2}{*}{1} & Sentence & 0.69 & 0.71 & 0.86 & 0.78 \\
                   & Study    & 0.77 & 0.73 & 0.97 & 0.83 \\ \hline
\multirow{2}{*}{2} & Sentence & 0.68 & 0.72 & 0.81 & 0.76 \\
                   & Study    & 0.75 & 0.72 & 0.94 & 0.82 \\ \hline
\multirow{2}{*}{3} & Sentence & 0.69 & 0.71 & 0.88 & 0.78 \\
                   & Study    & 0.72 & 0.69 & 0.97 & 0.80 \\ \hline
\multirow{2}{*}{4} & Sentence & 0.70 & 0.68 & 0.97 & 0.80 \\
                   & Study    & 0.65 & 0.63 & 1.00 & 0.77 \\ \hline
\multirow{2}{*}{5} & Sentence & 0.68 & 0.72 & 0.82 & 0.76 \\
                   & Study    & 0.77 & 0.73 & 0.97 & 0.83 \\ \hline
\textbf{AVG}       & Sentence & 0.69 & 0.71 & 0.87 & 0.78 \\
\textbf{AVG}       & Study    & 0.73 & 0.70 & 0.97 & 0.81 \\ \hline
\end{tabular}
}
\label{tab:scibert_scibert_results}
\end{table}

\subsection{BioBERT with \texttt{en\_core\_sci\_scibert}}
In this setup, we combined BioBERT with \texttt{en\_core\_sci\_scibert}. In the sentence-level the model achieved an accuracy of 0.69, precision 0.69, recall 0.93, and F1-score 0.79. At the study-level, accuracy was recorded as 0.72, precision averaged at 0.69, recall was 1.00, and F1-score was equal to 0.81. These findings show that recall was consistently high, while precision decreased slightly compared to the \texttt{en\_core\_sci\_md} variant, indicating that this configuration prefers sensitivity over specificity (see Table~\ref{tab:biobert_scibert_results}.

\begin{table}[htbp]
\centering
\caption{Performance of BioBERT Model \texttt{en\_core\_sci\_scibert} Across Five Attempts (Sentence- and Study-Level)}
\scalebox{0.85}
{
\begin{tabular}{c l c c c c}
\hline
\textbf{Attempt} & \textbf{Configuration} & \textbf{Accuracy} & \textbf{Precision} & \textbf{Recall} & \textbf{F1-score} \\ \hline
\multirow{2}{*}{1} & Sentence & 0.70 & 0.70 & 0.92 & 0.79 \\
                   & Study    & 0.75 & 0.71 & 1.00 & 0.83 \\ \hline
\multirow{2}{*}{2} & Sentence & 0.67 & 0.66 & 0.99 & 0.79 \\
                   & Study    & 0.60 & 0.60 & 1.00 & 0.75 \\ \hline
\multirow{2}{*}{3} & Sentence & 0.70 & 0.70 & 0.91 & 0.79 \\
                   & Study    & 0.75 & 0.71 & 1.00 & 0.83 \\ \hline
\multirow{2}{*}{4} & Sentence & 0.70 & 0.70 & 0.92 & 0.79 \\
                   & Study    & 0.75 & 0.71 & 1.00 & 0.83 \\ \hline
\multirow{2}{*}{5} & Sentence & 0.70 & 0.70 & 0.91 & 0.79 \\
                   & Study    & 0.75 & 0.71 & 1.00 & 0.83 \\ \hline
\textbf{AVG}       & Sentence & 0.69 & 0.69 & 0.93 & 0.79 \\
\textbf{AVG}       & Study    & 0.72 & 0.69 & 1.00 & 0.81 \\ \hline
\end{tabular}
}
\label{tab:biobert_scibert_results}
\end{table}

\subsection{Multiple Instance Learning with Entity Enriched}

Table~\ref{tab:mil_results} shows the averaged performance across five runs. The results present improvements in recall, with BioBERT achieving almost the highest F1-scores with different enrichment models. However, precision values remain modest, and recall approaches 1.0 across most configurations, reinforcing the sensitivity of entity-enriched MIL models for systematic review automation.

\begin{table}[htbp]
\centering
\caption{Performance of MIL with Entity-Enriched PLMs Across Different Configurations}
\scalebox{0.85}{
\begin{tabular}{l|l|c|c|c|c}
\hline
\textbf{Model} & \textbf{SpaCy Config}  & \textbf{Accuracy} & \textbf{Precision} & \textbf{Recall} & \textbf{F1-score} \\ \hline

\multirow{3}{*}{BERT} 
  & \multirow{1}{*}{en\_core\_sci\_sm}      & 0.62 & 0.61 & 1.00 & 0.76 \\
  \cline{2-6}
  & \multirow{1}{*}{en\_core\_sci\_md}    
                                              & 0.55 & 0.66 & 0.53 & 0.58 \\
  \cline{2-6}
  & \multirow{1}{*}{\textbf{en\_core\_sci\_scibert}}  & 0.65 & 0.63 & 1.00 & \textbf{0.77} \\
  \hline

\multirow{3}{*}{SciBERT} 
  & \multirow{1}{*}{en\_core\_sci\_sm}       & 0.63 & 0.63 & 0.92 & 0.75 \\
  \cline{2-6}
  & \multirow{1}{*}{\textbf{en\_core\_sci\_md}}      & 0.62 & 0.61 & 1.00 & \textbf{0.76} \\
  \cline{2-6}
  & \multirow{1}{*}{\textbf{en\_core\_sci\_scibert}}  & 0.63 & 0.63 & 0.97 & \textbf{0.76} \\
  \hline

\multirow{3}{*}{BioBERT} 
  & \multirow{1}{*}{en\_core\_sci\_sm}      & 0.62 & 0.61 & 1.00 & 0.76 \\
  \cline{2-6}
  & \multirow{1}{*}{\textbf{en\_core\_sci\_md}}       & 0.63 & 0.62 & 1.00 & \textbf{0.77} \\
  \cline{2-6}
  & \multirow{1}{*}{\textbf{en\_core\_sci\_scibert}}  & 0.65 & 0.63 & 1.00 & \textbf{0.77} \\
  \hline

\end{tabular}
}
\label{tab:mil_results}
\end{table}

\section{Discussion}
To evaluate the effectiveness of our proposed approaches, this section discusses the obtained results using both general-purpose (BERT) and domain-specific (SciBERT and BioBERT) PLMs, integrated with the domain-specific scispaCy pipeline (\texttt{en\_core\_sci\_sm}, \texttt{en\_core\_sci\_md}, and \texttt{en\_core\_sci\_scibert}), in addition to MIL approach with attention pooling.  We compare these results against three categories of baselines reported by Kertész et al.~\cite{kertesz2024eq5d}: (i)~classical bag-of-words models, (ii)~PLMs (BERT, SciBERT, BioBERT, and BlueBERT), and (iii)~fine-tuned PLMs (BERT, SciBERT, BioBERT, and BlueBERT) with classifier parameter tuning. Table~\ref{tab:comparison} presents the comparative results. Our experiments show that our proposed (fine-tuning with entity enrichment) pipeline significantly improves accuracy, precision, recall, and F1-score compared to the baseline models.

\begin{table}[htbp]
\centering
\caption{Summarized Performance Comparison (AVG across 5 attempts) vs. Baselines}
\scalebox{0.7}{
\begin{tabular}{ll|l|c|c|c|c}
\hline
\multicolumn{2}{c|}{\textbf{Baseline Models}} & \textbf{Dataset}  & \textbf{Accuracy} & \textbf{Precision} & \textbf{Recall} & \textbf{F1-score} \\ \hline

Naïve Bayes (BoW)  \hspace{23pt}     &  \hspace{0.45 in}       & Test set & 0.52 & 0.53 & 0.53 & 0.53 \\ 
Pretrained BERT    \hspace{23pt}      &   \hspace{0.45 in}     & Test set & N/A  & 0.68 & 0.63 & 0.62 \\ 
Fine-tuned BioBERT \hspace{23pt}      &    \hspace{0.45 in}   & Test set & 0.71 & 0.70 & 0.90 & 0.79 \\ \hline
\end{tabular}
}

\vspace{0.1cm}

\scalebox{0.7}{
\begin{tabular}{l|l|l|c|c|c|c}
\hline
\textbf{Model} & \textbf{SpaCy Config} & \textbf{Level} & \textbf{Accuracy} & \textbf{Precision} & \textbf{Recall} & \textbf{F1-score} \\ \hline

\multirow{6}{*}{BERT} 
  & \multirow{2}{*}{en\_core\_sci\_sm}      & Sentence & 0.69 & 0.70 & 0.92 & 0.79 \\
  &                                         & Study    & 0.69 & 0.67 & 0.99 & 0.80 \\ \cline{2-7}
  & \multirow{2}{*}{\textbf{en\_core\_sci\_md}}      & Sentence & 0.70 & 0.71 & 0.89 & 0.79 \\
  &                                         & \textbf{Study}    & \textbf{0.73} & \textbf{0.70} & \textbf{1.00} & \textbf{0.82} \\ \cline{2-7}
  & \multirow{2}{*}{en\_core\_sci\_scibert} & Sentence & 0.71 & 0.71 & 0.91 & 0.80 \\
  &                                         & Study    & 0.73 & 0.69 & 1.00 & 0.82 \\ \hline

\multirow{6}{*}{SciBERT} 
  & \multirow{2}{*}{en\_core\_sci\_sm}      & Sentence & 0.70 & 0.68 & 0.98 & 0.80 \\
  &                                         & Study    & 0.73 & 0.69 & 1.00 & 0.82 \\ \cline{2-7}
  & \multirow{2}{*}{en\_core\_sci\_md}      & Sentence & 0.70 & 0.70 & 0.93 & 0.80 \\
  &                                         & Study    & 0.72 & 0.68 & 1.00 & 0.81 \\ \cline{2-7}
  & \multirow{2}{*}{\textbf{en\_core\_sci\_scibert}} & Sentence & 0.69 & 0.71 & 0.87 & 0.78 \\
  &                                         & \textbf{Study}    & \textbf{0.73} & \textbf{0.70} & 0.97 & \textbf{0.81} \\ \hline

\multirow{6}{*}{BioBERT} 
  & \multirow{2}{*}{en\_core\_sci\_sm}      & Sentence & 0.69 & 0.69 & 0.93 & 0.79 \\
  &                                         & Study    & 0.74 & 0.70 & 0.99 & 0.82 \\ \cline{2-7}
  & \multirow{2}{*}{\textbf{en\_core\_sci\_md}}      & Sentence & 0.69 & 0.71 & 0.87 & 0.78 \\
  &                                         & \textbf{Study}    & \textbf{0.75} & \textbf{0.71} & 0.98 & \textbf{0.82} \\ \cline{2-7}
  & \multirow{2}{*}{en\_core\_sci\_scibert} & Sentence & 0.69 & 0.69 & 0.93 & 0.79 \\
  &                                         & Study    & 0.72 & 0.69 & 1.00 & 0.81 \\ \hline

  \multirow{1}{*}{BioBERT} 
  & \multirow{1}{*}{en\_core\_sci\_scibert}      & Bag of sentences & 0.65 & 0.63 & 1.00 & \textbf{0.77} \\
  \cline{1-7}

\end{tabular}
}
\label{tab:comparison}
\end{table}

\subsection{Comparison with Baseline Models}
\paragraph{Bag-of-Words based Naïve Bayes Classification}
Results from this approach indicate very good performance on the training dataset, but significantly lower performance on the testing dataset (F1-score 0.52 - 0.53), which indicates overfitting and, in turn, shows the ineffectiveness of shallow lexical features in determining the semantic details of scientific texts.

\paragraph{Pre-trained Language Models (PLMs)}
In this phase, Kertész et al.~\cite{kertesz2024eq5d} utilized four different PLMs (BERT, SciBERT, BioBERT, and BlueBERT) and achieved only moderate F1-scores ranging between 0.44 and 0.62 based on the architecture and input configuration.

\paragraph{Fine-tuning PLMs}
By fine-tuning PLMs (BERT, SciBERT, BioBERT, and BlueBERT) in different architectures and with different inputs, they performed much better, with the best BioBERT model achieving an average accuracy of 0.686.

\paragraph{Performance of Our Proposed Pipelines}
Our proposed pipelines demonstrates more stable and better generalization performances across both setups (sentence- and study- levels):
\begin{itemize}
    \item Sentence-level: average of F1-scores varied from 0.78 to 0.80. SciBERT with \texttt{en\_core\_sci\_sm} setup is able to achieve and maintain the best balances F1-score 0.80, and recall 0.98.
    \item Study-level: findings indicate that this approach consistently outperformed sentence-level, with F1-scores averaging between 0.81 and 0.82, with BioBERT with \texttt{en\_core\_sci\_md} achieving the highest F1-score of 0.82, and recall of 0.98.
    \item MIL (Bag of sentences): the best configuration in this approach is BioBERT with \texttt{en\_core\_sci\_scibert}, which able to reach accuracy 0.65, and F1-score 0.77.  
\end{itemize}

Notably, the achieved results in this study not only outperform the baseline, they also demonstrate robust generalization across both sentence- and study- level classification tasks. The recall values from our proposed pipeline are consistently near perfect at the study-level task, which indicates that the model correctly identifies relevant studies and maintains a high precision. This suggests a better generalization ability compared with the overfitting observed in baseline models.

\section{Conclusion and Future Directions}
This work proposes an entity-enriched fine-tuned pipeline, in addition to a MIL approach with attention pooling, that achieves robust performance across different configurations. The proposed approaches can achieve better F1-scores (reaching 0.82) and near-perfect recall at the study-level setting. It demonstrates substantial improvements in generalization and sensitivity, proving effective in detecting relevant EQ-5D studies when compared to baselines. Domain-specific PLMs (SciBERT, BioBERT) combined with scispaCy enrichment outperform general-purpose models, suggesting that integrating specialized biomedical knowledge is important. This study shows that entity enrichment is valuable for improving PLM effectiveness in SLR automation. There are several directions for future studies, such as:
\begin{itemize}
    \item Expand the dataset by utilizing semi-supervised approaches, which lead to better generalization, and perform the evaluation from 200 studies to thousands of abstracts. 
    \item Extend beyond binary detection of EQ-5D to other health-related quality-of-life instruments or clinical outcomes.
    \item Investigate if full-text enrichment results in better precision in detecting mentions of EQ-5D.
\end{itemize}

\section{Limitations}
In this section, we acknowledge several limitations of our study:
\begin{itemize}
    \item Limited generalization: The small size of the dataset (200 studies) may limit generalization, and model performance can vary with large, heterogeneous datasets.
    \item Language restriction: Considering only abstracts in English may reduce applicability for multilingual biomedical literature.
    \item Entity enrichment limitation: Entity enrichment is limited to scispaCy.
\end{itemize}

\section{Acknowledgement}

The authors would like to express their gratitude to the members of the Applied Machine Learning Research Group at Obuda University's John von Neumann Faculty of Informatics for their valuable comments and suggestions. They would also wish to acknowledge the support provided by the Doctoral School of Applied Informatics and Applied Mathematics at Obuda University.
\bibliographystyle{ieeetr}
\bibliography{ref}

\end{document}